\documentclass[]{ceurart}

\usepackage{amsmath}
\usepackage{amssymb}
\usepackage{tikz}
\usetikzlibrary{shapes.geometric, arrows, positioning}
\usepackage{booktabs}
\usepackage{array}

\begin{document}

\copyrightyear{2026}
\copyrightclause{for this paper by its authors. Use permitted under Creative Commons License Attribution 4.0 International (CC BY 4.0).}

\conference{CLEF 2026 Working Notes, 21--24 September 2026, Jena, Germany}

\title{BioSentinel at EXIST 2026: Soft-Label Optimization with XLM-RoBERTa for Sexism Intent Classification in Memes}

\author[1]{Chandru Munisamy}[%
orcid=0009-0008-3625-210X,
email=chandrumunisamy5@gmail.com,
]
\cormark[1]
\author[1]{Karthikeya Raguveer}[%
email=123ad0026@iiitk.ac.in,
]
\author[1]{Alapan Kuila}[%
email=alapan.cse@iiitk.ac.in,
]

\address[1]{Indian Institute of Information Technology, Design and Manufacturing, Kurnool, India}

\cortext[1]{Corresponding author.}

\begin{abstract}
This paper describes the BioSentinel team's participation in EXIST 2026 Task~2.2: Source Intention in Memes, part of the CLEF 2026 evaluation campaign.
The task requires classifying the communicative intent behind memes as \textsc{direct}, \textsc{judgemental}, or \textsc{no} (non-sexist), under a Learning with Disagreement (Le-Wi-Di) paradigm that mandates both hard-label and soft-label (probability distribution) predictions.
We present a text-centric approach built on \texttt{xlm-roberta-base} (270M parameters) trained with a composite loss function combining KL divergence on soft annotator distributions and weighted cross-entropy on hard labels.
On the official test set, the system achieved an ICM-Soft-Norm of $0.3229$ and ICM-Norm of $0.3778$, with a hard F1-score of $0.4236$, ranking 40th (out of 118 submissions) in the soft-soft evaluation and 49th (out of 187 submissions) in the hard-hard evaluation.
We provide an analysis of the dataset characteristics, exploratory larger-architecture runs, and the role of annotator disagreement in shaping model design for subjective NLP tasks. Ablation results show that KL loss improves soft-label metrics, while CE loss improves hard-label accuracy. We also report a separate validation-set temperature analysis.
\end{abstract}


\maketitle
\hypersetup{pageanchor=false}
\pagestyle{empty}
\thispagestyle{empty}
\pagenumbering{gobble}

\section{Introduction}
\label{sec:intro}

Online sexism~\cite{fox2015perpetuating} manifests in diverse forms across social media, ranging from overtly hostile statements to subtly coded memes that require cultural and visual context for interpretation~\cite{kirk2023semeval}.
The EXIST (sEXism Identification in Social neTworks) shared task series has established itself as a key benchmark for evaluating automated sexism detection systems~\cite{rodriguez2021overview,plaza2025overview}.
EXIST 2026 Task 2.2, \emph{Source Intention in Memes}, focuses on classifying the communicative intent of meme creators into three categories: \textsc{no} (not sexist), \textsc{direct} (explicit sexist intent), and \textsc{judgemental} (implicit or ironic sexist intent), as described in the official overview papers~\cite{plaza2026overview,plaza2026overview_extended}.

A defining feature of this task is the adoption of the Learning with Disagreement (Le-Wi-Di) paradigm~\cite{uma2021learning}. The complexity lies not just in identifying sexism, but in modeling the inherent subjectivity of human perception~\cite{davani2022dealing}. Annotators frequently disagree on what constitutes implicit sexism~\cite{rudman2000implicit} (e.g., the \textsc{judgemental} class). Consequently, models are required to output not just a single hard label, but a soft probability distribution representing the anticipated variance across human annotators~\cite{geng2016label}. This setting requires objectives that preserve soft annotator distributions rather than collapsing them to one-hot targets~\cite{song2022learning}.

Standard approaches to sexism detection typically collapse annotator disagreement~\cite{jiang2024re} into a majority-vote hard label and employ standard cross-entropy loss over pre-trained transformer models~\cite{mozafari2020hate}. While effective for objective tasks, this approach discards the valuable uncertainty signal present in subjective tasks and often leads to overconfident, poorly calibrated models.

We propose a text-centric approach built on \texttt{xlm-roberta-base} (270M parameters). Instead of relying purely on cross-entropy, we optimize a composite objective that jointly uses Kullback--Leibler (KL) divergence~\cite{kullback1951information} on the soft annotator distributions and weighted cross-entropy on the hard labels.

We deliberately focused on OCR text rather than full multimodal fusion for two reasons. First, OCR text was the most consistently aligned modality available for our training pipeline, whereas image and physiological signals would have required additional missing-feature and synchronization strategies. Second, our main objective was to isolate whether soft-label optimization could improve disagreement-aware prediction under a lightweight, reproducible bilingual setup. We therefore treat the absence of image modeling as a known limitation and examine its implications in the error analysis (Section~\ref{sec:error_analysis}).

Under our shared experimental configuration, larger architectures were less stable than \texttt{xlm-roberta-base}. These exploratory runs reused settings selected for the base model, so they do not establish a general conclusion about model scale. Prioritizing KL divergence in the loss is motivated by the goal of aligning predictions with annotator distributions, as examined in the ablation study (Section~\ref{sec:ablation}).

Our system\footnote{Implementation code is available at \url{https://github.com/kirito-meta/EXIST2026_PROJECT}.} produced soft-label predictions achieving an ICM-Soft-Norm of $0.3229$, an ICM-Norm of $0.3778$, and an F1 score of $0.4236$ on the official test set, ranking 40th in the soft-soft evaluation and 49th in the hard-hard evaluation. The system outperformed the majority-class hard baseline (ICM-Norm = $0.2375$, F1 score = $0.1676$) and the prior-distribution soft baseline (ICM-Soft-Norm = $0.2835$).

\section{Problem Definition}
\label{sec:problem}

\subsection{Task Formulation}
The core problem is to determine the intent behind a meme that may contain potentially sexist content. Because memes heavily rely on irony, juxtaposition, and cultural subtext, different people will interpret the same meme differently. The problem is twofold: predicting the most likely interpretation (hard classification) and predicting the spread of interpretations across a population (soft classification).

\subsection{Formal Problem Definition}
Although each benchmark instance contains an image $I$ and extracted OCR text $T$ in either English or Spanish, our system consumes OCR text only and learns a mapping function $f: T \rightarrow (y, P)$ where:
\begin{itemize}
  \item $y \in \{\textsc{no}, \textsc{direct}, \textsc{judgemental}\}$ is the hard predicted class.
  \item $P = [p_\textsc{no}, p_\textsc{direct}, p_\textsc{judgemental}]$ is a valid probability distribution such that $\sum p_i = 1.0$, representing the soft label prediction.
\end{itemize}

The task enforces a label hierarchy: $\mathcal{H} = \{\textsc{yes} \rightarrow [\textsc{direct}, \textsc{judgemental}], \; \textsc{no} \rightarrow []\}$. Confusing \textsc{no} with a sexist category incurs a higher penalty than confusing \textsc{direct} with \textsc{judgemental}.

\subsection{Dataset Description}
The dataset comprises 3,984 training instances and 1,053 test instances, balanced across English and Spanish. Each instance was evaluated by 6 human annotators.
The raw annotator votes were mapped to a soft gold probability distribution $p(c)$ for training. For example, an instance receiving 4 \textsc{direct}, 1 \textsc{judgemental}, and 1 \textsc{no} votes yields a soft label of $[0.167, 0.667, 0.167]$.

The official hard gold labels (assigned only if a class received strictly more than 2 votes out of 6) demonstrate severe class imbalance across the 3,149 resolvable instances: \textsc{no} accounts for 1,801 instances (57.2\%), \textsc{direct} for 902 instances (28.6\%), and \textsc{judgemental} for only 446 instances (14.2\%). The remaining 835 instances were highly ambiguous, receiving no majority class and thus excluded from the hard evaluation.
While the organizers also provided physiological sensor data (EEG, Eye Tracking, Heart Rate) from annotators, we ultimately excluded these due to a train-test feature mismatch (64 additional EEG features present in the test set but absent from training) and focused entirely on the OCR text modality.

\section{Proposed Method}
\label{sec:method}

\subsection{Architecture Overview}

\tikzstyle{process} = [rectangle, minimum width=3cm, minimum height=1cm, text centered, draw=black, fill=orange!10, rounded corners]
\tikzstyle{arrow} = [thick,->,>=stealth]
\tikzstyle{data} = [rectangle, minimum width=3cm, minimum height=1cm, text centered, draw=black, fill=blue!10]

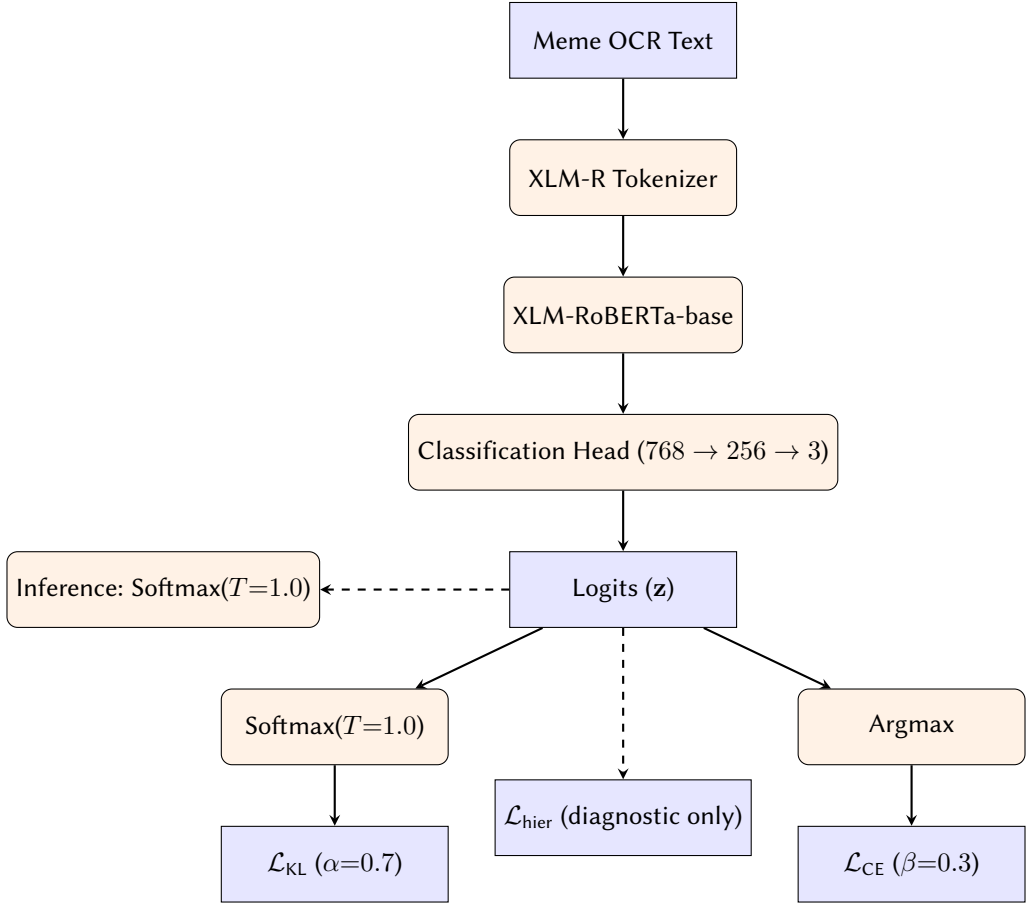
\begin{figure}[h]
\centering
\begin{tikzpicture}[node distance=0.8cm and 0.8cm]
\node (input) [data] {Meme OCR Text};
\node (tokenizer) [process, below=of input] {XLM-R Tokenizer};
\node (model) [process, below=of tokenizer] {XLM-RoBERTa-base};
\node (head) [process, below=of model] {Classification Head ($768 \rightarrow 256 \rightarrow 3$)};
\node (logits) [data, below=of head] {Logits ($\mathbf{z}$)};
\node (soft) [process, below left=of logits] {Softmax($T{=}1.0$)};
\node (hard) [process, below right=of logits] {Argmax};
\node (klloss) [data, below=of soft] {$\mathcal{L}_\text{KL}$ ($\alpha{=}0.7$)};
\node (celoss) [data, below=of hard] {$\mathcal{L}_\text{CE}$ ($\beta{=}0.3$)};
\node (hier) [data, below=2cm of logits] {$\mathcal{L}_\text{hier}$ (diagnostic only)};
\node (inference) [process, left=2.5cm of logits] {Inference: Softmax($T{=}1.0$)};

\draw [arrow] (input) -- (tokenizer);
\draw [arrow] (tokenizer) -- (model);
\draw [arrow] (model) -- (head);
\draw [arrow] (head) -- (logits);
\draw [arrow] (logits) -- (soft);
\draw [arrow] (logits) -- (hard);
\draw [arrow] (soft) -- (klloss);
\draw [arrow] (hard) -- (celoss);
\draw [arrow, dashed] (logits) -- (hier);
\draw [arrow, dashed] (logits) -- (inference);
\end{tikzpicture}
\caption{Architecture diagram. During training, $\mathcal{L}_\text{train} = \alpha \cdot \mathcal{L}_\text{KL} + \beta \cdot \mathcal{L}_\text{CE}$ updates weights. $\mathcal{L}_\text{hier}$ is logged as a non-differentiable diagnostic. At inference, probabilities are obtained by raw softmax ($T{=}1.0$).}
\label{fig:architecture}
\end{figure}

Our proposed methodology, illustrated in Figure~\ref{fig:architecture}, relies on a text-only pipeline using the \texttt{xlm-roberta-base} transformer model. We chose this model for its strong cross-lingual transfer capabilities, crucial for the bilingual nature of the dataset.

\subsection{Methodology Details}
\label{sec:methodology}

\paragraph{Classification Head.}
The first special-token representation (768-dimensional) from XLM-RoBERTa is passed through a multi-layer perceptron with dropout to produce 3-dimensional logits corresponding to the target classes.

\paragraph{Composite Loss Function.}
To address the Le-Wi-Di paradigm, the gradient-based training objective is:
\begin{equation}
  \label{eq:loss_train}
  \mathcal{L}_\text{train} = \alpha \cdot \mathcal{L}_\text{KL} + \beta \cdot \mathcal{L}_\text{CE},
\end{equation}
where $\alpha=0.7$ and $\beta=0.3$.

\begin{enumerate}
  \setlength{\itemsep}{0.30em}
  \setlength{\topsep}{0.25em}
  \setlength{\parsep}{0pt}
  \item \textbf{KL Divergence ($\mathcal{L}_\text{KL}$).}
  The primary objective minimizes the Kullback--Leibler divergence between the predicted distribution and the soft annotator distribution. This directly aligns the gradient-based training signal with the soft-label objective.

  \item \textbf{Weighted Cross-Entropy ($\mathcal{L}_\text{CE}$).}
  For instances with official hard labels, we use the fixed normalized class weights $[0.482, 0.572, 1.946]$ for \textsc{no}, \textsc{direct}, and \textsc{judgemental}, respectively, as implemented in the training script. These fixed weights increase the contribution of the under-represented \textsc{judgemental} class during training.

  \item \textbf{Hierarchy Boundary Diagnostic ($\mathcal{L}_\text{hier}$).}
  We also record a discrete boundary-error quantity when the hard prediction crosses the \textsc{no}/sexist boundary. It is computed from argmax predictions in a no-gradient block, so it does not update model parameters and was not used to select the reported checkpoint. The reported Epoch~4 checkpoint was selected solely by minimum validation KL divergence.
\end{enumerate}

\paragraph{Temperature Analysis.}
Official test predictions use raw softmax probabilities ($T=1.0$). We separately evaluated post-hoc temperature scaling on the held-out validation split. As shown in Section~\ref{sec:temperature}, $T=0.7$ improved validation ICM-Soft-Norm and reduced the reported validation ECE; this analysis was not used for the official test run.

\subsection{Model Training Parameters}
The system was implemented in PyTorch and trained on NVIDIA A100 GPUs via Google Colab Pro with the following hyperparameters:
\begin{itemize}
  \item \textbf{Base Model}: \texttt{FacebookAI/xlm-roberta-base} (270M parameters)
  \item \textbf{Hidden Layers (Head)}: $768 \rightarrow 256 \rightarrow 3$ (with GELU activation and 0.1 Dropout)
  \item \textbf{Epochs}: 6 (Epoch~4 checkpoint selected by minimum validation KL divergence)
  \item \textbf{Optimizer}: AdamW (weight decay = 0.01)
  \item \textbf{Learning Rate}: $2 \times 10^{-5}$ with linear warmup over 10\% of steps
  \item \textbf{Batch Size}: 32
  \item \textbf{Max Sequence Length}: 128 tokens
\end{itemize}

\section{Experiments and Results}
\label{sec:results}

\subsection{Validation and Official Test Results}
Our model was validated on a stratified 15\% hold-out validation set (598 instances), split by hard label and language to preserve the joint distribution of class proportions and bilingual balance. Table~\ref{tab:val_metrics} reports the validation analysis, including the post-hoc $T{=}0.7$ setting discussed in Section~\ref{sec:temperature}.

\begin{table}[htbp]
\caption{Validation set evaluation using PyEvALL.}
\label{tab:val_metrics}
\begin{tabular}{lr|lr}
\toprule
\multicolumn{2}{c}{\textbf{Hard-Hard Evaluation}} & \multicolumn{2}{c}{\textbf{Soft-Soft Evaluation}} \\
\midrule
ICM & $-0.482$ & ICM-Soft & $-1.670$ \\
ICM-Norm & 0.333 & ICM-Soft-Norm & 0.326 \\
F1 (macro) & 0.356 & Cross-Entropy & 1.468 \\
F1-\textsc{no} & 0.579 & Val KL Divergence & 0.352 \\
F1-\textsc{direct} & 0.418 & & \\
F1-\textsc{judgemental} & 0.071 & & \\
\bottomrule
\end{tabular}
\end{table}

The system's performance on the official EXIST 2026 test set is presented in Table~\ref{tab:test_metrics}.

\begin{table}[htbp]
\caption{Official test set results.}
\label{tab:test_metrics}
\begin{tabular}{lccc}
\toprule
\textbf{System} & \textbf{ICM-Soft-Norm} & \textbf{ICM-Norm} & \textbf{F1 (macro)} \\
\midrule
Majority-class baseline & --- & 0.2375 & 0.1676 \\
Prior-distribution baseline & 0.2835 & --- & --- \\
\textbf{BioSentinel (Ours)} & \textbf{0.3229} & \textbf{0.3778} & \textbf{0.4236} \\
\bottomrule
\end{tabular}
\end{table}

\subsection{Ablation Study}
\label{sec:ablation}

To understand the contribution of the gradient-based objectives, we conducted a single-split ablation study on the validation set. Table~\ref{tab:ablation} compares the loss configurations with the post-hoc temperature analysis.

\begin{table}[htbp]
\caption{Single-split ablation study on the validation set. All configurations use the same \texttt{xlm-roberta-base} architecture and hyperparameters. The final row reports a validation-only post-hoc analysis.}
\label{tab:ablation}
\begin{tabular}{lccc}
\toprule
\textbf{System configuration} & \textbf{ICM-Soft-Norm} & \textbf{ICM-Norm} & \textbf{F1 (macro)} \\
\midrule
CE only ($\beta{=}1.0$) & 0.142 & 0.285 & 0.348 \\
KL only ($\alpha{=}1.0$) & 0.312 & 0.310 & 0.294 \\
KL + CE & 0.315 & 0.321 & 0.354 \\
Selected KL + CE checkpoint, raw softmax ($T{=}1.0$) & 0.317 & 0.332 & 0.356 \\
Selected checkpoint + post-hoc $T{=}0.7$ (validation only) & 0.326 & 0.333 & 0.356 \\
\bottomrule
\end{tabular}
\end{table}

The ablation results support several observations. First, KL loss leads to better soft-label metrics (ICM-Soft-Norm of 0.312 vs.\ 0.142 under CE-only), whereas CE loss improves hard-label macro-F1 (0.348 vs.\ 0.294 under KL-only). Combining the objectives gives stronger joint development performance. For the selected checkpoint, post-hoc $T{=}0.7$ raises validation ICM-Soft-Norm from 0.317 to 0.326 while leaving the hard prediction unchanged.

\subsection{Validation Temperature Analysis}
\label{sec:temperature}

Table~\ref{tab:temperature} presents the temperature sweep on the validation set, showing ICM-Soft-Norm and Expected Calibration Error (ECE) at each temperature value.

\begin{table}[htbp]
\caption{Validation-set post-hoc temperature sweep.}
\label{tab:temperature}
\begin{tabular}{lcc}
\toprule
\textbf{Temperature} & \textbf{ICM-Soft-Norm} & \textbf{ECE} \\
\midrule
$T = 1.0$ (no scaling) & 0.317 & 0.185 \\
$T = 0.9$ & 0.321 & 0.159 \\
$T = 0.8$ & 0.324 & 0.138 \\
$T = 0.7$ (selected) & 0.326 & 0.118 \\
\bottomrule
\end{tabular}
\end{table}

Across the tested validation values, ICM-Soft-Norm increases from 0.317 at $T{=}1.0$ to 0.326 at $T{=}0.7$, while the reported validation ECE decreases from 0.185 to 0.118. This shows that sharpening improved the held-out validation measurements under this setup. These figures are from the validation split; official predictions follow the raw-softmax procedure described in Section~\ref{sec:methodology}. Temperature scaling remains a standard post-hoc adjustment~\cite{guo2017calibration}.

\subsection{Error Analysis}
\label{sec:error_analysis}

At the selected Epoch~4 checkpoint, 221 of the 598 validation instances had a raw hard prediction that differed from the official hard label. Table~\ref{tab:errors} summarizes recurring error mechanisms qualitatively; Table~\ref{tab:concrete_errors} provides three verified examples drawn from these errors.

\begin{table}[htbp]
\caption{Qualitative taxonomy used to interpret validation-set errors. These are descriptive mechanisms, not frequency estimates.}
\label{tab:errors}
\begin{tabular}{lp{9.6cm}}
\toprule
\textbf{Error Mechanism} & \textbf{Description} \\
\midrule
Sarcasm/Irony missed & OCR text may appear neutral or literal while framing changes the intended meaning. \\
Visual context missing & OCR text alone may be insufficient when a caricature, image, or visual juxtaposition carries the relevant signal. \\
OCR noise & Corrupted or incomplete OCR can omit words that determine the intended interpretation. \\
Cultural reference & Region-specific slang or cultural tropes may not be represented adequately by the text encoder. \\
\bottomrule
\end{tabular}
\end{table}

The qualitative taxonomy highlights irony and omitted visual context as recurring challenges for an OCR-only system. Because no controlled multimodal comparison was conducted, these observations are descriptive rather than causal. OCR noise and cultural references are additional plausible sources of error.

To make the error analysis concrete, Table~\ref{tab:concrete_errors} reports three genuine misclassified validation instances exported from the selected Epoch~4 checkpoint: one for each gold class. The soft distributions are ordered as $[\textsc{no}, \textsc{direct}, \textsc{judgemental}]$ and use raw softmax probabilities from the selected checkpoint. We provide OCR excerpts rather than reproducing the source images.

\begin{table}[!htbp]
\caption{Concrete validation-set misclassifications. Each row is a genuine held-out instance; OCR excerpts are shortened only for readability. Soft distributions are ordered as $[\textsc{no}, \textsc{direct}, \textsc{judgemental}]$.}
\label{tab:concrete_errors}
\footnotesize
\centering
\setlength{\tabcolsep}{3pt}
\begin{tabular}{>{\raggedright\arraybackslash}p{1.40cm}>{\raggedright\arraybackslash}p{2.65cm}>{\raggedright\arraybackslash}p{2.65cm}>{\raggedright\arraybackslash}p{5.95cm}}
\toprule
\textbf{ID / lang.} & \textbf{Gold / soft dist.} & \textbf{Pred. / soft dist.} & \textbf{OCR excerpt and observed error} \\
\midrule
\texttt{110627} (es) & \textsc{no}; [1.000, 0, 0] & \textsc{direct}; [0.221, 0.442, 0.336] & OCR (English translation): ``When you do not want to get up but remember that you are an empowered and independent woman\ldots'' The model assigns the highest probability to \textsc{direct} despite an all-\textsc{no} gold distribution. \\
\addlinespace
\texttt{210599} (en) & \textsc{direct}; [0.167, 0.667, 0.167] & \textsc{no}; [0.594, 0.259, 0.147] & OCR: ``I crave a relationship with traditional gender roles.'' The model predicts \textsc{no}, although the majority of annotator votes label its source intention \textsc{direct}. \\
\addlinespace
\texttt{210534} (en) & \textsc{judgemental}; [0.167, 0.167, 0.667] & \textsc{no}; [0.628, 0.197, 0.175] & OCR: ``Sitting at your family table when the biggest fish fry goes to your father and the smallest one sits on your mother's plate.'' The text-only system predicts \textsc{no}, while the majority gold distribution is \textsc{judgemental}. \\
\bottomrule
\end{tabular}
\end{table}

\subsection{Analysis and Discussion}

\paragraph{Soft-Label Alignment and Hard-Label Accuracy.}
The model's ICM-Soft-Norm of 0.3229 improves on the prior-distribution soft baseline (0.2835) by 13.9\% relative, whereas its hard macro-F1 is 0.4236. This pattern suggests that the model captures distributional signal not fully reflected by a single hard decision. The ablation study (Table~\ref{tab:ablation}) supports this interpretation: removing the KL term (CE-only) reduces ICM-Soft-Norm to 0.142, while CE remains important for hard-label accuracy.

\paragraph{The Challenge of the \textsc{judgemental} Class.}
The system performed poorly on the \textsc{judgemental} class (F1 = 0.071). Ironic and judgemental sexism can rely on the juxtaposition of text and image, while OCR text alone lacks contextual cues such as facial expressions and visual tropes. The qualitative taxonomy and the verified examples are consistent with missing visual context as a plausible contributor, but a controlled multimodal comparison would be required for causal attribution.

\paragraph{\textsc{judgemental} Cold-Start Behavior.}
Notably, the model produced zero \textsc{judgemental} predictions during the first two training epochs, with only 22 such predictions emerging at Epoch 3. At the selected Epoch~4 checkpoint, 43 out of 598 validation instances received a \textsc{judgemental} prediction. This cold-start behavior suggests that the model initially learns to distinguish only the \textsc{no}/\textsc{direct} boundary before gradually recognizing the more subtle \textsc{judgemental} patterns, indicating that longer training or dedicated \textsc{judgemental} augmentation strategies may be beneficial.

\paragraph{Architecture Scale and Stability.}
During development, we experimented with larger models (Gemma~2~2B, DeBERTa-v3-large, XLM-RoBERTa-large). Under our shared experimental configuration, these architectures were less stable than \texttt{xlm-roberta-base}: Gemma~2~2B exhibited divergent losses beyond epoch~2, DeBERTa-v3-large converged to majority-class predictions, and XLM-RoBERTa-large produced increasingly extreme probability distributions. Because these runs reused hyperparameters selected for the base model, they are exploratory rather than controlled architecture comparisons. A dedicated hyperparameter search would be required before drawing a general conclusion about model scale.

\section{Related Work}
\label{sec:related}

\paragraph{Sexism Detection in Social Media.}
Early approaches to online sexism detection relied on feature-engineered classifiers~\cite{talat2016hateful}, while recent methods leverage pre-trained transformer models fine-tuned on domain-specific data~\cite{devlin2019bert,conneau2020unsupervised}.
Multimodal meme classification has received increasing attention following the Hateful Memes Challenge~\cite{kiela2020hateful}, which underscored the need to reason jointly about text and images. Our work instead focuses on optimizing soft-label predictions from text alone and shows improvement over the reported soft baseline.

\paragraph{Learning with Disagreement.}
The Le-Wi-Di paradigm~\cite{uma2021learning} advocates preserving annotator disagreement rather than collapsing it into majority labels.
Soft-label training can preserve information about uncertainty in human labels~\cite{peterson2019human}.
The EXIST task series has progressively adopted this paradigm, making it a core evaluation criterion. Our composite loss design is directly informed by this paradigm, using KL divergence as the primary training signal to preserve annotator distribution information.

\paragraph{Model Calibration.}
Neural networks are known to be poorly calibrated~\cite{guo2017calibration}. Temperature scaling, where logits are divided by a scalar $T$ before softmax, is a standard post-hoc adjustment. Our validation analysis found lower reported ECE and higher ICM-Soft-Norm at $T{=}0.7$ (Table~\ref{tab:temperature}).

\paragraph{Evaluation with ICM.}
The Information Contrast Model~\cite{amigo2022evaluating} provides a hierarchy-aware evaluation metric that penalizes cross-level confusions more severely than within-level errors. It remains useful for interpreting cross-boundary mistakes in the hard-label evaluation.

\section{Limitations and Ethics}
\label{sec:limitations}

Our study has several limitations. First, the text-only approach inherently cannot capture visual-textual interactions that can be central to meme interpretation, which is especially relevant to the low \textsc{judgemental}-class F1. The qualitative error taxonomy identifies irony and omitted visual context as recurring patterns, but does not establish causal contribution without a controlled multimodal comparison. Second, with only 3,984 training instances, the generalizability of our findings to broader meme datasets remains uncertain. Third, the ablation study was conducted on a single validation split; multi-seed experiments would provide more robust estimates of each component's contribution. Fourth, sexism detection systems should not be deployed for automated content moderation without human oversight, as false positives could lead to unjust censorship and false negatives could allow harmful content to propagate. Finally, the training data may contain cultural and linguistic biases that are reflected in the model's predictions, particularly given the bilingual but limited geographic scope of the annotator pool.

\section{Conclusion}
\label{sec:conclusion}

In this shared-task setting, the BioSentinel \texttt{xlm-roberta-base} system combines KL divergence with weighted cross-entropy to model annotator disagreement from OCR text. On the official test set, it achieves an ICM-Soft-Norm of 0.3229.

Our ablation study shows that KL loss improves the soft-label metric, while CE loss improves hard-label accuracy; combining them ($\alpha{=}0.7, \beta{=}0.3$) balances both objectives. The post-hoc validation analysis at $T{=}0.7$ improves ICM-Soft-Norm by 0.009 and lowers the reported validation ECE. Temperature scaling was assessed only on validation data; official predictions use raw softmax. The hierarchy boundary error is used only for diagnostic analysis.

Our exploratory runs suggest that stability under noisy soft labels warrants further study. Future work should integrate visual features to address the weakness on \textsc{judgemental} sexism, evaluate components across multiple random seeds, and investigate ways to align physiological signals with textual and visual inputs.

\begin{acknowledgments}
We thank the EXIST 2026 organizers for providing the dataset, evaluation framework, and the opportunity to participate.
We also acknowledge Google Colab for providing the computational infrastructure used in our experiments.
\end{acknowledgments}

\begin{aideclaration}
During the preparation of this work, the authors used ChatGPT (GPT-4o) and Gemini for grammar and spelling checks, text restructuring, and code debugging. The authors reviewed and edited all generated suggestions and take full responsibility for the final content.
\end{aideclaration}



\begingroup\small
\bibliography{reference}
\endgroup
\end{document}